\documentclass[letterpaper, 10 pt, conference]{ieeeconf}  

\IEEEoverridecommandlockouts                              

\usepackage{times}
\usepackage{epsfig}
\usepackage{graphicx}
\usepackage{amsmath}
\usepackage{amssymb}

\usepackage{booktabs}
\usepackage{multirow}
\newcommand\mdoubleplus{\mathbin{+\mkern-10mu+}}

\title{\LARGE \bf
HGCN-GJS: Hierarchical Graph Convolutional Network with Groupwise Joint Sampling for Trajectory Prediction
}

\author{Yuying Chen$^{*}$, Congcong Liu$^{*}$, Xiaodong Mei$^{*}$, Bertram E. Shi and Ming Liu
\thanks{*This work was supported by Zhongshan Science and Technology Bureau Fund, under project 2020AG002, Foshan-HKUST Project no. FSUST20-SHCIRI06C, and Guangdong Basic and Applied Basic Research Foundation project no. 2020A0505090008, awarded to Prof. Ming Liu.(\textit{$^{*}$The authors contributed equally to this work. Corresponding author: Ming Liu})}
\thanks{Yuying Chen, Congcong Liu, Xiaodong Mei and Bertram E. shi are with The HongKong University of Science and Technology, Hong Kong SAR, China. Ming Liu is with The HongKong University of Science and Technology (Guangzhou), Nansha, Guangzhou, 511400, Guangdong, China, The HongKong University of Science and Technology, Hong Kong SAR, China, and HKUST Shenzhen-Hong Kong Collaborative Innovation Research Institute, Futian, Shenzhen. Email:
        {\tt\small (ychenco,cliubh,xmeiab)@connect.ust.hk;
        (eebert,eelium)@ust.hk}}%
}

\begin{document}

\maketitle
\thispagestyle{empty}
\pagestyle{empty}

\begin{abstract}

Pedestrian trajectory prediction is of great importance for downstream tasks, such as autonomous driving and mobile robot navigation.
Realistic models of the social interactions within the crowd is crucial for accurate pedestrian trajectory prediction.
However, most existing methods do not capture group level interactions well, focusing only on pairwise interactions and neglecting group-wise interactions.
In this work, we propose a hierarchical graph convolutional network, HGCN-GJS, for trajectory prediction which well leverages group level interactions within the crowd.
Furthermore, we introduce a joint sampling scheme that captures co-dependencies between pedestrian trajectories during trajectory generation. 
Based on group information, this scheme ensures that generated trajectories within each group are consistent with each other, but enables different groups to act more independently. We demonstrate that our proposed network achieves state of the art performance on all datasets we have considered.

\end{abstract}

\section{INTRODUCTION}


Predicting human trajectories is crucial for many tasks including crowd aware robot navigation, crowd surveillance systems, and autonomous driving.
One of the main challenges of making accurate trajectory prediction is to model the complex social interactions within the crowd
, which are mainly driven by social rules and social relationships. For example,  when walking in a crowd, pedestrians will avoid passing through a group of people talking to each other. People tend to follow coherent trajectories with friends and maintain a comfortable distance from others. 

\begin{figure}[ht!] 
  \centering
  \includegraphics[width=0.85\columnwidth]{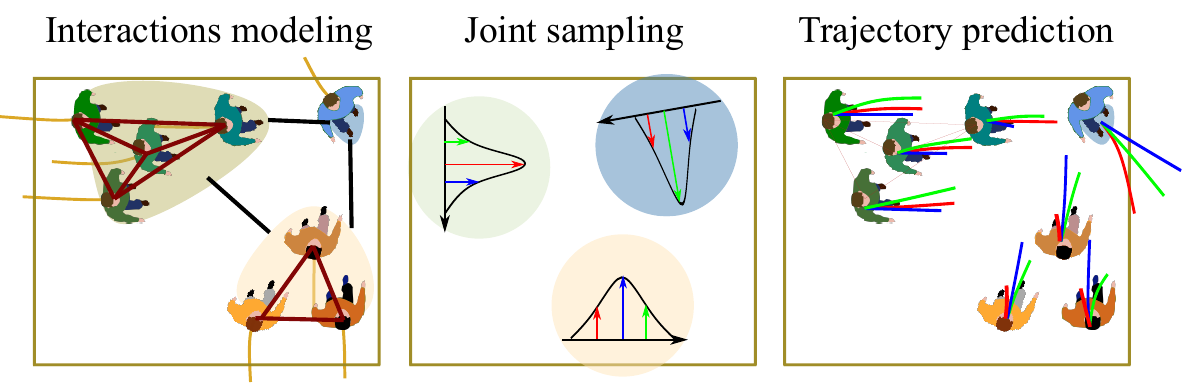}
  \caption{Pedestrian trajectory prediction that fully utilizes the group information. A single hierarchical GCN captures the intragroup and intergroup interactions for all pedestrians. Joint sampling of the latent space is conducted based on the group constraints. This results in joint predictions with more aligned and realistic trajectories.      \label{fig:idea} }
  \vspace{-1em}
\end{figure}

According to Moussaid \textit{et al.} \cite{moussaid2010walking}, up to 70\% of pedestrians are moving in groups.
However, group level interactions within the crowd have been rarely considered. How group members interact with each other, with other pedestrians and with other groups has not been systematically investigated. 
Recent work introduced motion coherence as a way to set adjacency in graph-based interaction modeling \cite{chen2020comogcn}. 
However, 
group labels were only used to modulate pairwise interactions and higher level social interactions between groups was neglected.
In this work, we propose a hierarchical graph convolutional network to model the interactions between and within the groups better.
The lower level of the hierarchy models interactions between individual pedestrians within groups. 
The upper level of the hierarchy models interactions between groups.

Furthermore, in most previous work, which either adopts a GAN \cite{gupta2018social, sadeghian2019sophie, amirian2019social} or a VAE \cite{chen2020comogcn, ivanovic2019trajectron, mangalam2020not} as backbone network to model the stochasticity of trajectory prediction, predictor has sampled trajectories either independently for each pedestrian or globally across the entire scene.
However, sampling trajectories independently per pedestrian can cause unrealistic predictions, such as collisions between members of the same group. 
On the other hand, sampling the same latent vector for all pedestrians in the scene introduces too much coupling. 
To address these shortcomings, we propose a joint sampling scheme for generating correlated predictions within groups. 

As illustrated in Fig. \ref{fig:idea},  through the hierarchical GCN and joint sampling operation, we fully capture group information.
Our main contributions are as follows: 
\begin{itemize}
\item We propose a novel hierarchical graph representation that captures intra and inter group behavior at different levels of the hierarchy.

\item We propose a joint sampling scheme with group constraints for generating consistent trajectories for pedestrians within groups.

\item With above benefits, the proposed HGCN-GJS
achieves state-of-the-art performance on multiple trajectory prediction benchmarks.



\end{itemize}

\section{RELATED WORK}

\subsection{Interactions Modeling for Trajectory Prediction}



The ways to handle social interactions in recent trajectory prediction works can be roughly classified into pooling based methods and graph based methods.
Social LSTM \cite{alahi2016social} used the social pooling, which pools the hidden states of neighbors according to the distance from pedestrian of interest. 
Social GAN \cite{gupta2018social} used a max-pooling module to aggregate information from pedestrians in the crowd, but only captured pairwise interactions.
SoPhie \cite{sadeghian2019sophie} adopted a soft attention operation to integrate information across pedestrians and about the physical scene.
More recent work has used graphs to model interactions, e.g. GCNs \cite{chen2020comogcn, chen2020robot,liu2021avgcn}, GATs \cite{huang2019stgat,kosaraju2019social} and message passing neural networks \cite{hu2020collaborative}.
The use of multi-head attention in GAT increases its computational complexity compared to GCN. 
In this work, we utilize a GCN to integrate interaction between humans and between groups.

\subsection{Group Information for Trajectory Prediction}
Motion coherency within the crowd contains implicit but rich information of social relationships.
Several approaches have been proposed to detect coherent motion within the crowd.
Zhou \textit{et al.} \cite{zhou2012coherent} introduced coherent filtering methods for motion coherency detection which work well for crowds with large crowd densities. 
Based on it, Chen \textit{et al.} proposed a hybrid labeling method leveraging DBSACN clustering method to compensate for the drawbacks of coherent filtering in small group detection \cite{chen2020comogcn}. 

To date, only a few works have considered group interactions for trajectory prediction. 
Group-LSTM \cite{bisagno2018group} selectively included information about outsiders during the social pooling process.
Sun \textit{et al.} \cite{sun2020recursive} learned a relation matrix to distinguish whether two pedestrians are in the same group, and applied this matrix to a GCN for interaction modeling. They did not separate intergroup or intragroup interactions. 
Chen \textit{et al.} better exploited the group information by using two GCNs: one to tackle the interactions within the group and the other to handle interactions with pedestrians outside the group \cite{chen2020comogcn}.  
In this work, we propose a novel hierarchical graph representation that better captures interactions within and between groups to improve interaction modeling.

\subsection{Sampling Scheme for Stochastic prediction}
For generative models, a sampling procedure is conducted to introduce stochasticity. 
The decoding stages of previous RNN-based works \cite{chen2020comogcn} sampled trajectories independently for each pedestrian. 
However, it is not reasonable for pedestrians in the same group, whose trajectories are coupled.
In this work, we propose a novel joint sampling scheme to generate joint behaviors that are more realistic.

\section{METHODOLOGY}

\subsection{Problem Definition}
Our goal is to predict future trajectories of pedestrians based on their observed past trajectories. We assume there are $N$ pedestrians distributed in $M$ groups in the scene during an observed time window $({t_1}$: $t_{\text{obs}})$. 
The observed trajectories are denoted as $x_{1,...,N}^{(t_1:t_{\text{obs}})} = \{(\text{x}_{1,...,N}^{t_1},\text{y}_{1,...,N}^{t_1}), ..., (\text{x}_{1,...,N}^{t_\text{obs}},\text{y}_{1,...,N}^{t_\text{obs}})\}$, where $(\text{x} ,\text{y})$ are the coordinates of the pedestrians. 
Most works simply represent the trajectories as $x_{\text{rel}_{1,...,N}}^{(t_1:t_{\text{obs}})}$, where $x_{\text{rel}_{i}}^t$ denotes the position of pedestrian $i$ at time step $t$ relative to the position at $t-1$. 
The relative trajectory $x_{\text{rel}_{1,...,N}}^{(t_1:t_{\text{obs}})}$ reveals the motion pattern of pedestrians, but it does not contain the relative spatial information of pedestrians. 
We use $p_{\text{rel}_i}^{(1:N)}$ , which denotes the positions of pedestrians $1,...,N$ relative to pedestrian $i$ at $t_{\text{obs}}$, to capture this information.
The predicted trajectories over $T$ time steps is defined as $\hat{x}_{\text{rel}_{1,...,N}}^{(t_{\text{obs}}:t_{\text{obs}}+T)}$.

\subsection{Overall Model}
\begin{figure*}[htb]
  \centering
  \includegraphics[width=0.85\textwidth]{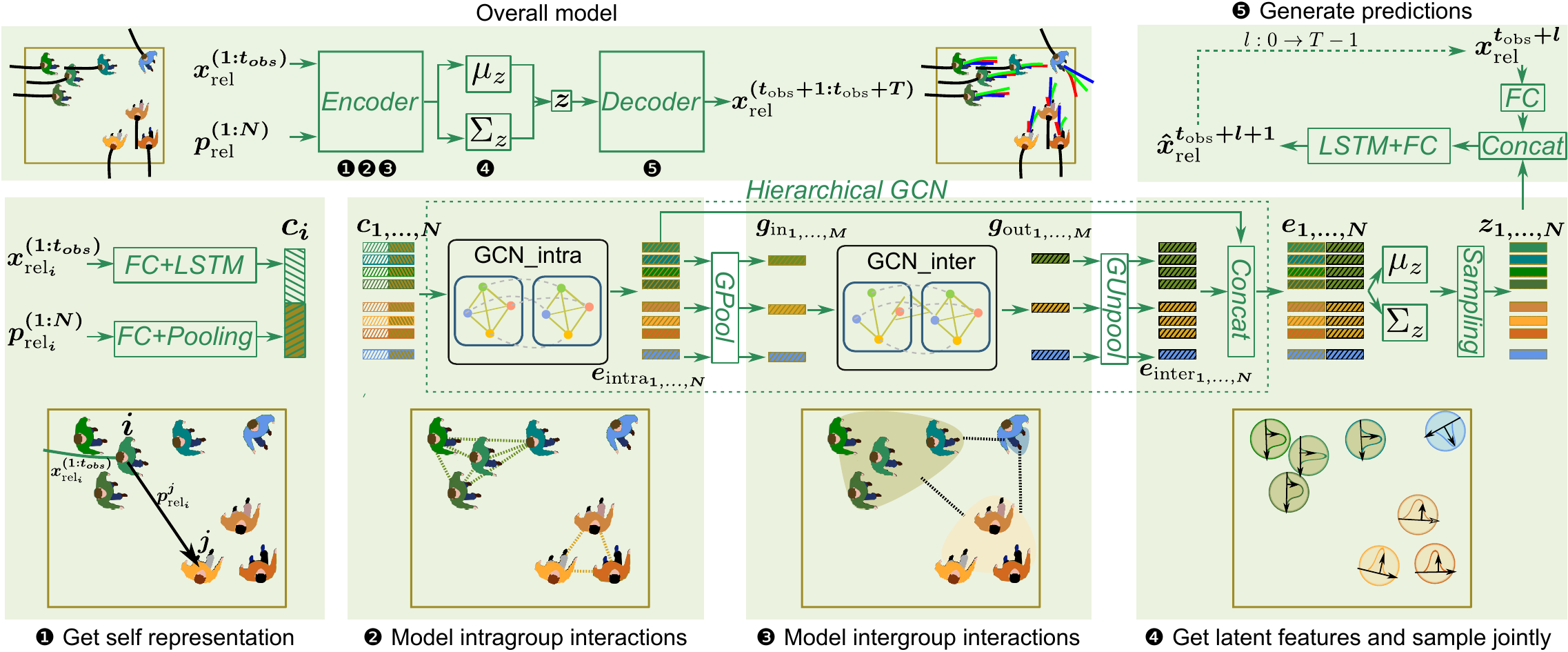}
  \caption{The variational trajectory prediction network. The network takes the past trajectory as input and predicts multiple future trajectories. The modeling process can be split into 5 steps: 1. For each pedestrian, we obtain the self representation that includes the self motion pattern and context. 2. Intragroup interactions are modeled with \textbf{GCN\_intra}, which constructs fully connected graphs within the groups. 3. After the \textbf{GPool} process for each group, \textbf{GCN\_inter} takes the group features as input and construct a fully connected graph for intergroup interactions modeling. 4. After the \textbf{GUnpool} process, the group-wise intergroup interaction features are concatenated with the intragroup features. The resulting embedding is used for creating a distribution with mean $\mu$ and variance $\Sigma$. Joint sampling process is applied and samples inside a group are designed to have large positive correlation coefficients (corr), while samples from different groups are designed to have corr close to zero. 5. The sampled features are fed into a decoder LSTM to generate the predicted trajectories.\label{fig:structure} }
  \vspace{-1em}
\end{figure*}

Overall model follows a variational encoder-decoder structure. As shown in the upper left of Fig. \ref{fig:structure}, it consists of an encoder that encodes the past trajectories into features, a sampler that generates latent codes $\boldsymbol{z}$ following the distribution $\boldsymbol{z\sim\mathcal{N}(\mu,\Sigma)}$, and a decoder that generates trajectory predictions. The whole process can be split into five steps.

First, we obtain the self representation for each pedestrian that encodes the motion information ($x_{\text{rel}}^{(t_1:t_{\text{obs}})}$) and relative spatial information ($p_{\text{rel}}^{(1:N)}$). 
We first use two single-layer MLPs (FCs) to obtain the embeddings. Then an LSTM is utilized to extract the motion features and average pooling is conducted to combine the spatial features. These features are concatenated ($\mdoubleplus$) to get $c_{i}$.
\begin{equation}
\label{eqn:self}
c_i=\text{LSTM}_{\text{en}}(\text{MLP}_{\text{mot}}(x_{\text{rel}_i}^{(t_1:t_\text{obs})})) \mdoubleplus \text{Avg.}(\text{MLP}_{\text{sp}}(p_{\text{rel}_i}^{(1:N)}))
\end{equation}
Second, we build a graph that regards each pedestrian as a node, and use a two-layer graph convolutional network (GCN), \textbf{GCN\_intra}, to model the intragroup interactions: 
\begin{equation}
e_{\text{intra}_{1,...,N}} = \text{GCN}_{\text{intra}}(c_{1,...,N}, A_{\text{intra}})
\end{equation}
Third, the generated features ($e_{\text{intra}_{1,...,N}}$) go through a group-wise pooling \textbf{GPool} module to obtain the representation for each group ($g_{\text{in}_{1,...,M}}$). We build a graph that treats each group as a node, and use a two-layer GCN (\textbf{GCN\_inter}) to model the intergroup interactions. The generated intergroup features ($g_{\text{out}_{1,...,M}}$) then go through a \textbf{GUnpool} module to be distributed to each pedestrian. 
\begin{gather}
g_{\text{in}_{1,...,M}} = \text{GPool}(e_{\text{intra}_{1,...,N}}) \\
g_{\text{out}_{1,...,M}} = \text{GCN}_{\text{inter}}(g_{\text{in}_{1,...,M}}, A_{\text{inter}})\\
e_{\text{inter}_{1,...,N}} = \text{GUnpool}(g_{\text{out}_{1,...,M}})
\end{gather}
Fourth, the generated intragroup ($e_{\text{intra}_{1,...,N}}$) and intergroup ($e_{\text{inter}_{1,...,N}}$) features for each pedestrian are concatenated as $e_{1,...,N}$, which is further used to generate mean and variance vector of the distribution $\boldsymbol{z\sim\mathcal{N}(\mu,\Sigma)}$ through two MLPs. A joint sampling process generates vectors $\boldsymbol{z}$ that exhibit positive correlations among members of the same group. 
\begin{gather}
e_{1,...,N} = e_{\text{intra}_{1,...,N}} \mdoubleplus e_{\text{inter}_{1,...,N}} \\ 
\mu_{1,...,N}, \Sigma_{1,...,N} = \text{MLP}_{\mu_z}(e_{1,...,N}), \text{MLP}_{\Sigma_z}(e_{1,...,N}) \\
z_{1,...,N} = \text{Sampling}(\mu_{{1,...,N}}, \Sigma_{{1,...,N}})
\end{gather}
Fifth, the sampled latent features are concatenated with the embedding computed from the last predicted state and then together fed into an LSTM. The output of the LSTM goes through an FC to get the predicted next state. This process loops $T$ times to generate the predicted trajectories.
\begin{gather}
   x_{\text{rel}}^{t_\text{obs}+l+1} =  \text{MLP}(\text{LSTM}(z_{1,...,N} \mdoubleplus \text{MLP}_\text{de}(x_{\text{rel}}^{t_\text{obs}+l})))
\end{gather}


\subsection{Hierarchical GCN for Interaction Modeling}
\begin{figure}[t]
  \centering
  \includegraphics[width=0.7\columnwidth]{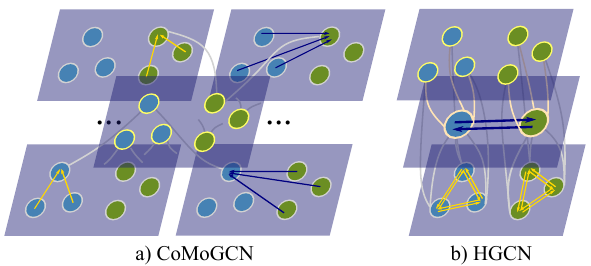}
  \caption{Compared with the flat structure of the CoMoGCN and other group-based trajectory prediction methods, the hierarchical GCN (HGCN) models interactions at different levels of the hierarchy. 
  \label{fig:hgcn} }
\vspace{-1.2em}
\end{figure}

As an efficient and straightforward method for interactions modeling, the GCN has been widely used for trajectory prediction. 
However, previous work mainly considered only pairwise human-human interactions by constructing multiple star-topology graphs. CoMoGCN \cite{chen2020comogcn} considered group interactions by two parallel GCNs for each pedestrian: one for pairwise intragroup interactions and one for pairwise extragroup interactions, which neglects group level interactions.
As depicted in Fig. \ref{fig:hgcn}, our work introduces two new ways of modeling group information. 
First, for intergroup or intragroup interactions modeling, rather than the $2N$ GCNs used in \cite{chen2020comogcn}, we used one GCN to account for all the pedestrians. 
This reduced the amount of computation from $2N^2$ to $N+M$.
Second, we propose a hierarchical method that first models the intragroup interactions (individual level) and then models the intergroup interactions (group level). 
The method is inspired by considering the pedestrians and groups as analogous to atoms and molecules, where intramolecular forces within the molecule bind together the atoms and intermolecular forces mediate interactions between molecules. 

For intragroup interactions modeling, we represent each person as the node and establish fully connected graph inside each group, as shown in Fig. \ref{fig:hgcn}(b). The adjacency matrix $A_{\text{intra}}$ is calculated from the coherency mask $M_{\text{intra}}$ accordingly, where both $A_{\text{intra}}$ and $M_{\text{intra}} \in \mathbb{R}^{N\times N}$. $M_{\text{intra}}$ indicates whether two pedestrians are in the same group:
\begin{gather} 
M_{\text{intra}}(i,j) = \left\{\begin{matrix}
1, & \text{if person $i$,$j$ in the same group or $i=j$}    \\ 
0, & \text{else}
\end{matrix}\right.
\\
A_{\text{intra}} = \text{Normalize}(M_{\text{intra}})
\end{gather}
The normalization ensures each row sums to 1.

Entries of $M_{\text{intra}}$ are determined by group labels, which are obtained by hybrid coherent motion clustering method proposed in \cite{chen2020comogcn}. First, pedestrians are classified into multiple groups roughly with the coherent filtering based on the correlation intensity between their relative positions and velocities, as computed from the trajectory histories. Then the DBSCAN method is utilized to refine the group labels with the consideration of motion directions and relative angular distances, in order to detect small groups. 

To get the representative feature for each group, we deploy a group-wise average pooling process. We first obtain the $M$ unique rows from $M_{\text{intra}}$, which results in $R_{\text{intra}} \in \mathbb{R}^{M \times N}$ and then normalize. Thus, the \textbf{GPool} process is represented as: 
\begin{gather}
R_{\text{intra}} = \text{Unique}(M_{\text{intra}}) \\
g_{\text{in}_{1,...,M}} = \text{Normalize}(R_{\text{intra}})\cdot e_{\text{intra}_{1,...,N}}
\end{gather}
For intergroup interactions modeling, we consider group-wise interactions and establish a fully connected graph of all groups, as shown in the middle layer in Fig. \ref{fig:hgcn}(b). The adjacency matrix $A_{\text{inter}} \in \mathbb{R}^{M \times M}$ is normalized row-wise:
\begin{equation}
A_{\text{inter}} = \text{Normalize}(M_{\text{inter}}),
\end{equation}
where $M_{\text{inter}} \in \mathbb{R}^{M \times M}$ is an all-ones matrix, which indicates that all groups have equal influence on each other. 

To distribute the group-wise features to all pedestrians, we deploy a \textbf{GUnpool} process:
\begin{equation}
\label{eqn:gunpool}
e_{\text{inter}_{1,...,N}} =  R_{\text{intra}}^\text{T} \cdot g_{\text{out}_{1,...,M}} 
\end{equation}

The intragroup and intergroup features are concatenated to give the features of each pedestrian for further processing, as shown in the top layer of the structure in Fig. \ref{fig:hgcn}(b). 

\subsection{Joint Sampling for Coherent Trajectories}
\begin{figure}[htb]
  \centering
  \includegraphics[width=0.5\columnwidth]{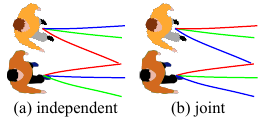}
  \caption{Predicted multiple trajectories by different sampling strategy.
  \label{fig:sampling} }
\end{figure}

Previous works generated trajectories independently, by sampling latent code from independent marginal distribution.
This often results in unrealistic trajectories. As shown in Fig. \ref{fig:sampling}(a), independent sampling may generate trajectories that violate motion coherency, by predicting crossed trajectories (red).
To address it, we introduce the joint sampling. 

Rather than sampling from $z \sim \mathcal{N}(\mu,\Sigma)$, in practice, a reparameterization trick is adopted in the variational autoencoder (VAE). 
For each pedestrian $i$, we obtain $z_i$ by
\begin{equation}
z_{i} = \mu_{i} + \Sigma^{1/2}_{i} \cdot \epsilon_{i}
\end{equation}
For independent sampling, each $\epsilon_i \in \mathbb{R}^{D}$ is sampled independently, where $D$ is the size of the latent code. To incorporate the motion coherence into the sampling process, we introduce correlations between the $\epsilon_i$ vectors for pedestrians in the same group. If we concatenate the $\epsilon_i$ vectors into a single $\epsilon \in \mathbb{R}^{N \times D}$ vector, then we sample this vector from $\epsilon \sim \mathcal{N}(0, \Sigma_g)$, where
\begin{equation}
\label{eqn:corr}
    \Sigma_{g} = 
\begin{bmatrix}
 1  & \rho_{12} &\cdots&  \rho_{1N} \\
  \rho_{21} & 1   & \cdots &\rho_{2N} \\
    \vdots &\vdots & \ddots & \vdots\\
  \rho_{N1} & \rho_{N2} & \cdots & 1\\ 
\end{bmatrix} \otimes I_{D},
\end{equation}
where $\otimes$ indicates the Kronecker product, $I_{D}$ is the $D\times D$ identity matrix and $\rho_{ij} = \rho$ if person $i$ and $j$ are in the same group and zero otherwise. If $\rho = 1$, then we can obtain $\epsilon$ from the concatenation of $M$ independent sampled vectors, $\epsilon_{g}\in \mathbb{R}^{M\times D}$, using the \textbf{GUpool} process:

\begin{equation}
\epsilon = R_{\text{intra}}^\text{T} \cdot \epsilon_{g},
\end{equation}
where each $\epsilon_{g_i}$ in $\epsilon_{g}$ is sampled from $\epsilon_{g_i} \sim \mathcal{N}(0, I_D)$. 


\section{EXPERIMENTS}

\subsection{Implementation details}
$\text{MLP}_{\text{mot}}$, $\text{MLP}_{\text{sp}}$, and $\text{MLP}_{\text{de}}$ had a single fully conneted layer with output dimension 16. The hidden dimension of $\text{LSTM}_{\text{en}}$ was 32. The two GCNs had two layers with dimensions 72 and 16. $\text{MLP}_{\mu_z}$ and $\text{MLP}_{\Sigma_z}$ had a single FC layer with output dimension 8. 
$\text{LSTM}_{\text{de}}$ had hidden dimension 32. $\text{MLP}$ had output dimension 2.
We used ReLU as the activation function. 

\subsection{Training and evaluation}
\subsubsection{Dataset for training and evaluation}
The training and evaluations are conducted on pedestrian trajectory datasets: ETH \cite{pellegrini2009you} and UCY \cite{lerner2007crowds} with the coherent motion labels provided by \cite{chen2020comogcn}.
The ETH dataset consists of two sets: ETH and Hotel. 
The UCY dataset consists of three sets: Univ, Zara1, and Zara2. 
There are five sets with four different scenes and 1536 pedestrians in total. 
We followed SGAN's data loader, taking trajectories with 8 time steps as observations and evaluating predictions over the next 12 time steps.
\subsubsection{Loss and other settings used for training}
The models were trained for 400 epochs with the Adam optimizer.
The mini-batch size was 16 and the learning rate was 1e-4. 
The loss function contains two parts, shown in (\ref{eqn:loss}). The first part is the L1 distance between the estimated trajectory and the ground truth trajectory. The second part is the Kullback–Leibler divergence from $q$ to $z$, where $q \sim \mathcal{N}(0,1)$.  
\begin{equation}
\label{eqn:loss}
   L_{\text{pred}} = ||\boldsymbol{x_{\text{rel}_i}} - \boldsymbol{\hat{x}_{\text{rel}_i}}||+ \alpha KL(\boldsymbol{z}, \boldsymbol{q}).
\end{equation}
Following SGAN, we adopted the variety loss with k=20. During training, 20 trajectories are predicted for each pedestrian and the one with minimum loss for parameter update.


\subsubsection{Evaluation Metrics}
We use two standard metrics for evaluations:
(a) Average Displacement Error (\textit{ADE}):  Average L2 distance in meters over all time steps between the ground truth and the predicted trajectory;
(b) Final Displacement Error (\textit{FDE}): Average L2 distance in meters at the final time step between the ground truth and the predicted trajectory.

\subsubsection{Baselines for comparison}
We compare our model with the following recent works based on generative methods:
(i) \textit{Social GAN (SGAN)} \cite{gupta2018social}: A generative model leveraging GAN to generate stochastic predictions, with a global max-pooling module to integrate crowd interactions;
(ii) \textit{SoPhie} \cite{sadeghian2019sophie}: A GAN based model, which also considers physical interaction with environmental constraints;
(iii) \textit{Trajectron} \cite{ivanovic2019trajectron}: A generative model using the CVAE for stochastic predictions with a spatiotemporal graph structure.;
(iv) \textit{Social-BiGAT} \cite{kosaraju2019social}: A generative model using a Bicycle-GAN for stochastic prediction and a GAT for crowd interaction modeling;
(v) \textit{CoMoGCN} \cite{chen2020comogcn}: A generative model using a GCN for crowd interaction aggregation and incorporating  coherent motion information;
(vi) \textit{STGAT} \cite{huang2019stgat}: A generative model using a spatial-temporal graph structure;
(vii) \textit{NMMP} \cite{hu2020collaborative}: A generative model using neural message passing module. 

\begin{figure*}[htb]
\centering
\includegraphics[width=0.75\textwidth, trim=120 115 0 0, clip]{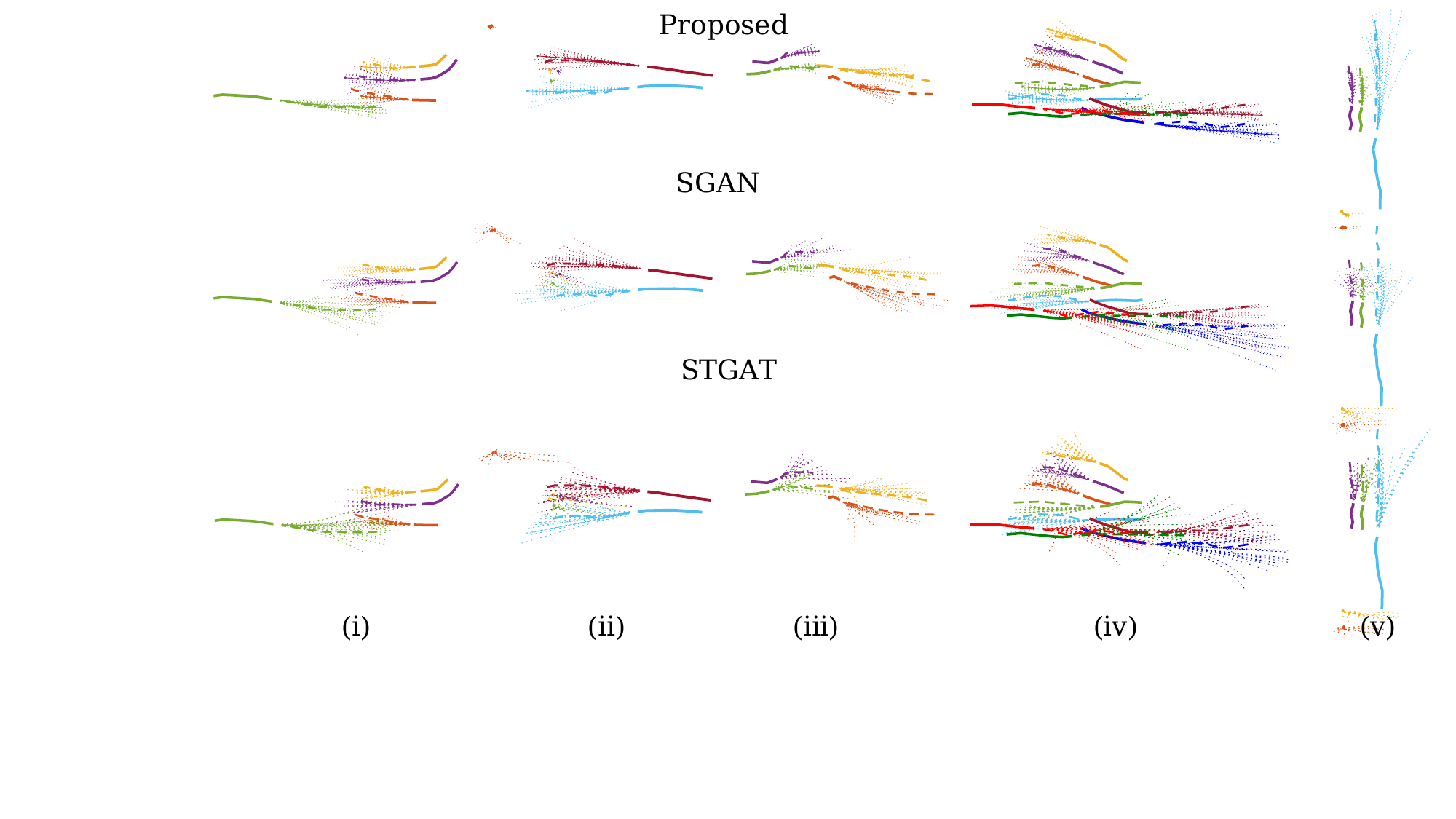}
\caption{HGCN-GJS compared with SGAN and STGAT. The 20 samples generated are shown in thin dashed lines. The observed past trajectories are shown in solid lines. The ground truth future trajectories are shown in wide dashed lines. The predictions generated by the mean value ($\mu$) are shown as dot-dashed lines. Different colors denote different pedestrians.\label{fig:visual_sota} }
\end{figure*}


\begin{table*}[]
\caption{
Quantitative results of trajectory prediction for comparisons. (lower value denotes better performance).
}
\vspace{-1em}
\centering
\begin{tabular}{|l|ccccccc|c|l|l}
\hline
        & \multicolumn{7}{c|}{Baselines}                                                                                                                                                                      & \multicolumn{3}{c|}{Ours}        \\ \cline{2-11} 
Dataset & \multicolumn{1}{c|}{SGAN} & \multicolumn{1}{c|}{Sophie} & \multicolumn{1}{c|}{Trajectron} & \multicolumn{1}{c|}{S-BiGAT} & \multicolumn{1}{c|}{CoMoGCN} & \multicolumn{1}{c|}{STGAT} & NMMP        & \multicolumn{3}{c|}{HGCN-GJS}       \\ \hline
ETH     & 0.81/1.52                  & 0.70/1.43                   & 0.59/1.14                       & 0.69/1.29                    & 0.70/1.28                    & 0.70/1.21                  & 0.67/1.22   & \multicolumn{3}{c|}{\textbf{0.58/1.01}}   \\
HOTEL   & 0.72/1.61                  & 0.76/1.67                   & 0.35/0.66                       & 0.49/1.01                    & 0.37/0.75                    & 0.32/0.63                  & 0.33/0.64   & \multicolumn{3}{c|}{\textbf{0.26/0.47}}   \\
UNIV    & 0.60/1.26                  & 0.54/1.24                   & 0.54/1.13                       & 0.55/1.32                    & 0.53/1.16                    & 0.56/1.20                  & \textbf{0.52/1.12}   & \multicolumn{3}{c|}{0.53/1.16}   \\
ZARA1   & 0.34/0.69                  & 0.30/0.63                   & 0.43/0.83                       & \textbf{0.30/0.62}                    & 0.34/0.71                    & 0.33/0.64                  & 0.32/0.66   & \multicolumn{3}{c|}{0.32/0.65}   \\
ZARA2   & 0.42/0.84                  & 0.38/0.78                   & 0.43/0.85                       & 0.36/0.75                    & 0.31/0.67                    & 0.30/0.61                  & 0.29/0.62   & \multicolumn{3}{c|}{\textbf{0.25/0.53}}   \\ \hline
AVG     & 0.58/1.18                  & 0.54/1.15                   & 0.47/0.92                      & 0.48/1.00                    & 0.45/0.91                    & 0.44/0.86                  & 0.43/0.85 & \multicolumn{3}{c|}{\textbf{0.39/0.76}} \\ \hline
\end{tabular}
\label{tab:pred_error_baseline}
\vspace{-1em}
\end{table*}

\section{Results}
\subsection{Comparison with state-of-the-art methods}
Table \ref{tab:pred_error_baseline} compares our model and the previous models. 
The ADE and FDE were listed across five datasets.
Our proposed HGCN-GJS significantly outperforms all baselines by 9.3\%-32.8\% on ADE and 10.6\%- 35.6\% on FDE on average.

Fig. \ref{fig:visual_sota} shows several examples of the generated trajectories from our model, SGAN and STGAT. Trajectories generated by our method are distributed more accurately around the ground truth trajectories (the trajectories in (i), (ii) and (iii)). Trajectories predicted by the proposed model also show lower variance for pedestrians moving in groups (shown in (iv)), due to the constraints introduced by the coherent motion. 
This is particularly noticeable for still pedestrians (e.g. the still pedestrians in the groups in (ii) (v)). 
We observed that both the SGAN and the STGAT generated lower or higher amplitude motions, while our proposed method gives trajectories that are more consistent with the ground truth (shown in (ii) and (v)). 

\subsection{Ablation study}
\subsubsection{Hierarchical GCN vs. Parallel GCN}
\label{sec:hiervspara}
\begin{figure}[t]
  \centering
  \includegraphics[width=0.7\columnwidth]{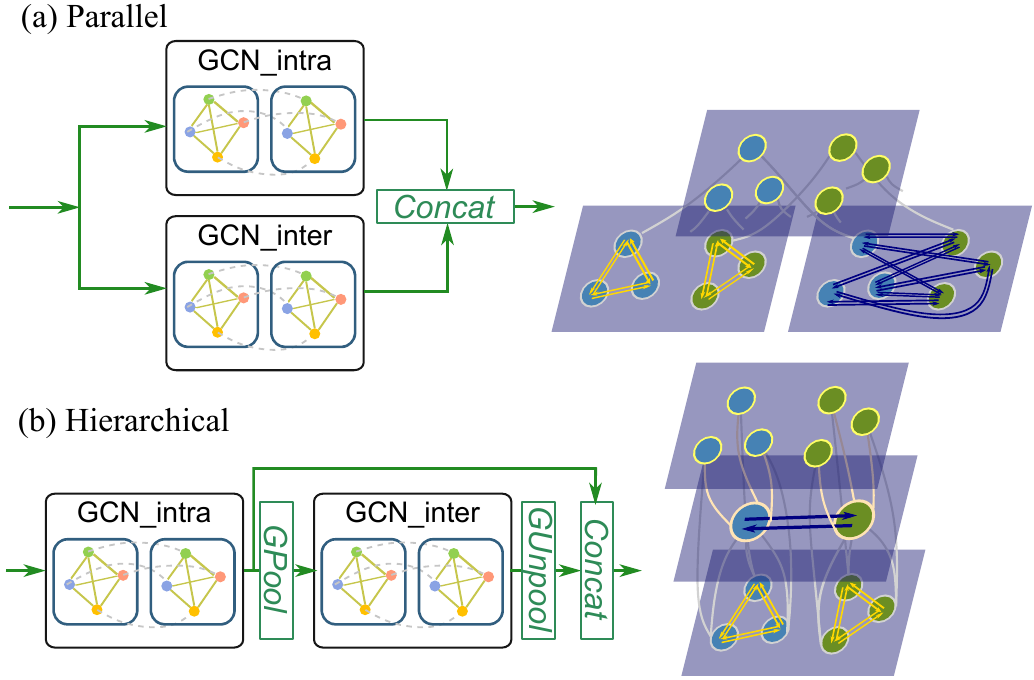}
  \caption{Ablation study to show the advantage of the hierarchical design. \label{fig:hgcn_abla} }
  \vspace{-0.5em}
\end{figure}

\begin{table}[t]
\caption{Performance comparison of different structure designs.}
\vspace{-1em}
 \resizebox{\columnwidth}{!}{

\begin{tabular}{|l|llccccc|}
\hline
                                                                          &     & ETH  & HOTEL & UNIV & ZARA1 & ZARA2 & AVG   \\ \hline
\multirow{2}{*}{\begin{tabular}[c]{@{}l@{}}Para-\\ llel\end{tabular}}     & ADE & 0.73 & 0.33  & 0.53 & 0.34  & 0.26  & 0.44 \\
                                                                          & FDE & 1.45 & 0.68  & 1.16 & 0.71  & 0.56  & 0.91 \\ \hline
\multirow{2}{*}{\begin{tabular}[c]{@{}l@{}}Hierar-\\ chical\end{tabular}} & ADE & 0.73 & 0.30  & 0.52 & 0.33  & 0.27  & 0.43 \\
                                                                          & FDE & 1.35 & 0.59  & 1.13 & 0.69  & 0.57  & 0.87 \\ \hline
\end{tabular}}

\label{tab:structure}
\vspace{-1.5em}
\end{table}
To show the advantage of the hierarchical GCN design, we implemented a similar model where the intragroup and intergroup information was integrated by two GCNs placed in parallel. As shown in Fig. \ref{fig:hgcn_abla}, it used only 2 GCNs: one for intragroup and one for extragroup pairwise interactions, rather than 2N GCNs (two per pedestrian).
For both models, we used independent sampling across pedestrians for the consistency.
As shown in Table \ref{tab:structure}, using the hierarchical GCN design improves ADE by 1.8\% and FDE by 5.0\% on average, which demonstrates the better performance of the hierarchical design. 
We also note that the performance of the parallel model is similar to that of the CoMoGCN, indicating that there is no advantage to using separate GCNs for each pedestrians.

\subsubsection{Joint sampling vs. independent sampling}

\begin{figure}[t]
\centering
\includegraphics[width=0.55\columnwidth]{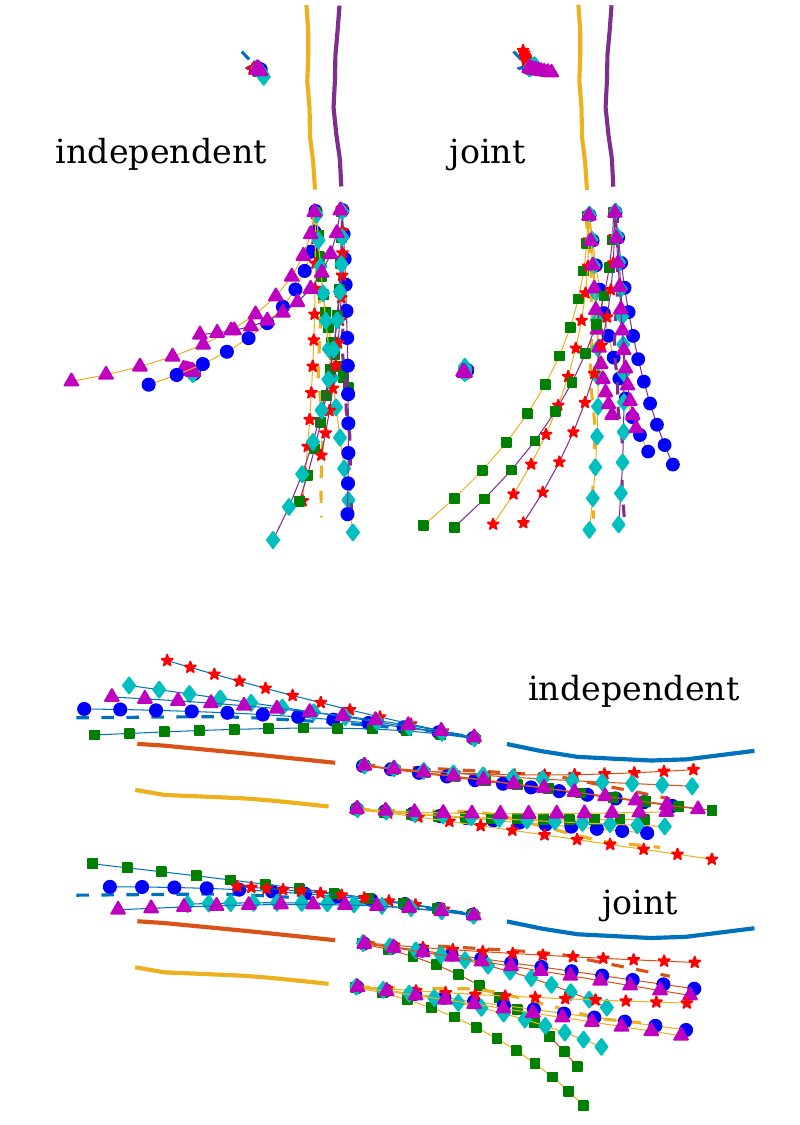}
\caption{Independent sampling vs. joint sampling. Trajectories shown with different markers are trajectories generated at different times. The observed past trajectories are shown in solid lines, ground truth future trajectories are shown in wide dashed lines. \label{fig:sampling_result} }
\vspace{-1.5em}
\end{figure}

We compared trajectory samples generated by joint sampling and independent sampling. Fig. \ref{fig:sampling_result} shows representative examples for two pedestrians travelling together. 
In the upper pair of examples, the model with independent sampling generates trajectories that often diverge (blue) or intersect (cyan and green). Similar results can be observed in the bottom pair of examples. 
Trajectories generated by joint sampling tend to remain side-by-side and maintain the same speed.

Our proposed joint sampling within groups lies as at an intermediate point between independent sampling for all pedestrians and a single latent variable
which introduces coupling between the trajectories of groups that are very far apart from each other. 
To compare, we further calculated the collision rates for different methods. As shown in Table \ref{tab:collison-rate}, HGCN-GJS (joint sampling within groups) achieves the lowest rate compared with HGCN (independent sampling) and SGAN (joint sampling among all pedestrians).

\begin{table}[t]
\caption{Collision rate of the predicted trajectories (as the multiple of the collision rate of HGCN-GJS).}
\vspace{-1em}
\resizebox{\columnwidth}{!}{
\begin{tabular}{|l|cccccc|}
\hline
                               & \multicolumn{1}{c|}{ETH} & \multicolumn{1}{c|}{HOTEL} & \multicolumn{1}{c|}{UNIV} & \multicolumn{1}{c|}{ZARA1} & \multicolumn{1}{c|}{ZARA2} & AVG \\ \hline
\multicolumn{1}{|c|}{HGCN-GJS} & 1.0                      & 1.0                        & 1.0                       & 1.0                        & 1.0                        & 1.0 \\ \cline{1-1}
HGCN                           & 2.0                      & 1.4                        & 1.3                       & 3.3                        & 1.7                        & 1.9 \\ \cline{1-1}
SGAN                           & 1.2                      & 1.1                        & 1.6                       & 0.9                        & 1.1                        & 1.2 \\ \hline
\end{tabular}}

\label{tab:collison-rate}
\end{table}

\begin{figure}[t]
  \centering
  \includegraphics[width=0.6\columnwidth]{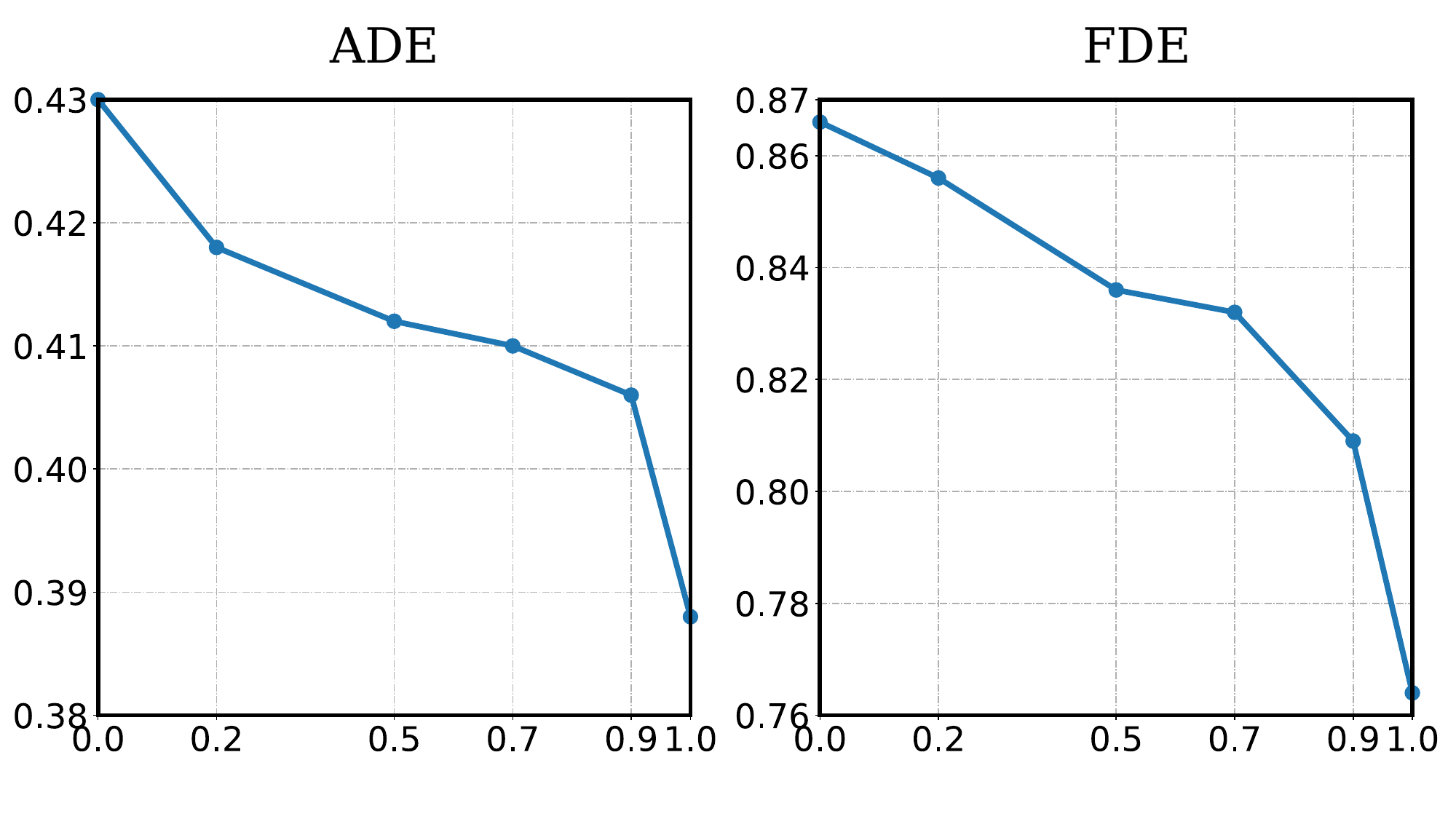}
  \caption{Average ADE and FDE with different $\rho$ settings. \label{fig:corr} }
  \vspace{-2em}
\end{figure}

To show the influence of the correlation within each group, we set the $\rho$ in Eqn. (\ref{eqn:corr}) to 0, 0.2, 0.5, 0.7, 0.9, and 1 respectively. 
The average ADE and FDE of models trained with different $\rho$ settings are shown in Fig. \ref{fig:corr}. 
With the increase of $\rho$, both ADE and FDE decrease.
Compared with the model with independent sampling, the average ADE and FDE of our proposed model is 9.3\% and 12.6\% smaller. 

\section{CONCLUSIONS}

In this paper, we propose a novel generative model, HGCN-GJS for trajectory prediction.
We design a hierarchical graph convolutional network for group level interaction aggregation within the crowd. 
Our hierarchical network enables us to focus on pairwise interactions within groups, and to model the interactions between groups at a higher level of abstraction.
Furthermore, we introduce a novel joint sampling scheme for generating stochastic trajectories in a more realistic and reasonable way.
Our results demonstrate that each innovation leads to significant improvements in the performance of trajectory prediction, and that when combined together, the benefits compound. 










\bibliographystyle{IEEEtran}
\bibliography{iros.bib}

\begin{thebibliography}{10}
\providecommand{\url}[1]{#1}
\csname url@rmstyle\endcsname
\providecommand{\newblock}{\relax}
\providecommand{\bibinfo}[2]{#2}
\providecommand\BIBentrySTDinterwordspacing{\spaceskip=0pt\relax}
\providecommand\BIBentryALTinterwordstretchfactor{4}
\providecommand\BIBentryALTinterwordspacing{\spaceskip=\fontdimen2\font plus
\BIBentryALTinterwordstretchfactor\fontdimen3\font minus
  \fontdimen4\font\relax}
\providecommand\BIBforeignlanguage[2]{{%
\expandafter\ifx\csname l@#1\endcsname\relax
\typeout{** WARNING: IEEEtran.bst: No hyphenation pattern has been}%
\typeout{** loaded for the language `#1'. Using the pattern for}%
\typeout{** the default language instead.}%
\else
\language=\csname l@#1\endcsname
\fi
#2}}

\bibitem{moussaid2010walking}
M.~Moussa{\"\i}d, N.~Perozo, S.~Garnier, D.~Helbing, and G.~Theraulaz, ``The
  walking behaviour of pedestrian social groups and its impact on crowd
  dynamics,'' \emph{PloS one}, vol.~5, no.~4, p. e10047, 2010.

\bibitem{chen2020comogcn}
Y.~Chen, C.~Liu, B.~Shi, and M.~Liu, ``Comogcn: Coherent motion aware
  trajectory prediction with graph representation,'' \emph{The 31st British
  Machine Vision Virtual Conference (BMVC 2020)}, 2020.

\bibitem{gupta2018social}
A.~Gupta, J.~Johnson, L.~Fei-Fei, S.~Savarese, and A.~Alahi, ``Social {GAN}:
  Socially acceptable trajectories with generative adversarial networks,'' in
  \emph{Proceedings of the IEEE Conference on Computer Vision and Pattern
  Recognition}, 2018, pp. 2255--2264.

\bibitem{sadeghian2019sophie}
A.~Sadeghian, V.~Kosaraju, A.~Sadeghian, N.~Hirose, H.~Rezatofighi, and
  S.~Savarese, ``So{P}hie: An attentive {GAN} for predicting paths compliant to
  social and physical constraints,'' in \emph{Proceedings of the IEEE
  Conference on Computer Vision and Pattern Recognition}, 2019, pp. 1349--1358.

\bibitem{amirian2019social}
J.~Amirian, J.-B. Hayet, and J.~Pettr{\'e}, ``Social ways: Learning multi-modal
  distributions of pedestrian trajectories with {GAN}s,'' in \emph{Proceedings
  of the IEEE Conference on Computer Vision and Pattern Recognition Workshops},
  2019, pp. 0--0.

\bibitem{ivanovic2019trajectron}
B.~Ivanovic and M.~Pavone, ``The {T}rajectron: Probabilistic multi-agent
  trajectory modeling with dynamic spatiotemporal graphs,'' in
  \emph{Proceedings of the IEEE International Conference on Computer Vision},
  2019, pp. 2375--2384.

\bibitem{mangalam2020not}
K.~Mangalam, H.~Girase, S.~Agarwal, K.-H. Lee, E.~Adeli, J.~Malik, and
  A.~Gaidon, ``It is not the journey but the destination: Endpoint conditioned
  trajectory prediction,'' in \emph{European Conference on Computer
  Vision}.\hskip 1em plus 0.5em minus 0.4em\relax Springer, 2020, pp. 759--776.

\bibitem{alahi2016social}
A.~Alahi, K.~Goel, V.~Ramanathan, A.~Robicquet, L.~Fei-Fei, and S.~Savarese,
  ``Social {LSTM}: Human trajectory prediction in crowded spaces,'' in
  \emph{Proceedings of the IEEE Conference on Computer Vision and Pattern
  Recognition}, 2016, pp. 961--971.

\bibitem{chen2020robot}
Y.~Chen, C.~Liu, B.~E. Shi, and M.~Liu, ``Robot navigation in crowds by graph
  convolutional networks with attention learned from human gaze,'' \emph{IEEE
  Robotics and Automation Letters}, vol.~5, no.~2, pp. 2754--2761, 2020.

\bibitem{liu2021avgcn}
C.~Liu, Y.~Chen, M.~Liu, and B.~E. Shi, ``Avgcn: Trajectory prediction using
  graph convolutional networks guided by human attention,'' in \emph{2021 IEEE
  International Conference on Robotics and Automation (ICRA)}.\hskip 1em plus
  0.5em minus 0.4em\relax IEEE, 2021, pp. 14\,234--14\,240.

\bibitem{huang2019stgat}
Y.~Huang, H.~Bi, Z.~Li, T.~Mao, and Z.~Wang, ``{STGAT}: Modeling
  spatial-temporal interactions for human trajectory prediction,'' in
  \emph{Proceedings of the IEEE International Conference on Computer Vision},
  2019, pp. 6272--6281.

\bibitem{kosaraju2019social}
V.~Kosaraju, A.~Sadeghian, R.~Mart{\'\i}n-Mart{\'\i}n, I.~Reid, H.~Rezatofighi,
  and S.~Savarese, ``Social-{B}i{GAT}: Multimodal trajectory forecasting using
  {B}icycle-{GAN} and graph attention networks,'' in \emph{Advances in Neural
  Information Processing Systems}, 2019, pp. 137--146.

\bibitem{hu2020collaborative}
Y.~Hu, S.~Chen, Y.~Zhang, and X.~Gu, ``Collaborative motion prediction via
  neural motion message passing,'' in \emph{Proceedings of the IEEE/CVF
  Conference on Computer Vision and Pattern Recognition}, 2020, pp. 6319--6328.

\bibitem{zhou2012coherent}
B.~Zhou, X.~Tang, and X.~Wang, ``Coherent filtering: Detecting coherent motions
  from crowd clutters,'' in \emph{European Conference on Computer
  Vision}.\hskip 1em plus 0.5em minus 0.4em\relax Springer, 2012, pp. 857--871.

\bibitem{bisagno2018group}
N.~Bisagno, B.~Zhang, and N.~Conci, ``Group {LSTM}: Group trajectory prediction
  in crowded scenarios,'' in \emph{The European Conference on Computer Vision
  Workshops}, September 2018.

\bibitem{sun2020recursive}
J.~Sun, Q.~Jiang, and C.~Lu, ``Recursive social behavior graph for trajectory
  prediction,'' in \emph{Proceedings of the IEEE/CVF Conference on Computer
  Vision and Pattern Recognition}, 2020, pp. 660--669.

\bibitem{pellegrini2009you}
S.~Pellegrini, A.~Ess, K.~Schindler, and L.~Van~Gool, ``You'll never walk
  alone: Modeling social behavior for multi-target tracking,'' in \emph{2009
  IEEE 12th International Conference on Computer Vision}.\hskip 1em plus 0.5em
  minus 0.4em\relax IEEE, 2009, pp. 261--268.

\bibitem{lerner2007crowds}
A.~Lerner, Y.~Chrysanthou, and D.~Lischinski, ``Crowds by example,'' in
  \emph{Computer Graphics Forum}, vol.~26, no.~3.\hskip 1em plus 0.5em minus
  0.4em\relax Wiley Online Library, 2007, pp. 655--664.

\end{thebibliography}

\end{document}